\documentclass[10pt,letterpaper]{article}
\usepackage[top=0.85in,left=2.75in,footskip=0.75in]{geometry}

\usepackage{amsmath,amssymb}

\usepackage{changepage}

\usepackage[utf8x]{inputenc}

\usepackage{textcomp,marvosym}

\usepackage{cite}

\usepackage{nameref,hyperref}

\usepackage[right]{lineno}

\usepackage{microtype}
\DisableLigatures[f]{encoding = *, family = * }

\usepackage[table]{xcolor}

\usepackage{array}

\newcolumntype{+}{!{\vrule width 2pt}}

\newlength\savedwidth

\newcommand\thickhline{\noalign{\global\savedwidth\arrayrulewidth\global\arrayrulewidth 2pt}%
\hline
\noalign{\global\arrayrulewidth\savedwidth}}

\usepackage{verbatimbox}

\raggedright
\usepackage[aboveskip=1pt,labelfont=bf,labelsep=period,justification=raggedright,singlelinecheck=off]{caption}

\makeatletter
\renewcommand{\@biblabel}[1]{\quad#1.}
\makeatother

\usepackage{lastpage,fancyhdr,graphicx}
\usepackage{epstopdf}
\fancyheadoffset[L]{2.25in}
\fancyfootoffset[L]{2.25in}
\begin{document}
\vspace*{0.2in}

\begin{flushleft}
{\Large
\textbf\newline{\textbf{GPT Takes the Bar Exam}} 
}
\newline
\\
Michael Bommarito II \textsuperscript{1,2,3},
Daniel Martin Katz\textsuperscript{1,2,3,*}
\\
\bigskip
\textbf{1} Illinois Tech - Chicago Kent College of Law (Chicago, IL  USA)
\\
\textbf{2} Bucerius Law School (Hamburg, Germany)
\\
\textbf{3} CodeX - The Stanford Center for Legal Informatics (Stanford, CA  USA)
\\
\bigskip

%
%



* Corresponding Author: dkatz3@kentlaw.iit.edu

\end{flushleft}
\section*{Abstract}
Nearly all jurisdictions in the United States require a professional license exam, commonly referred to as ``the Bar Exam,'' as a precondition for law practice.  To even sit for the exam, most jurisdictions require that an applicant completes at least seven years of post-secondary education, including three years at an accredited law school. In addition, most test-takers also undergo weeks to months of further, exam-specific preparation.  Despite this significant investment of time and capital, approximately one in five test-takers still score under the rate required to pass the exam on their first try.  In the face of a complex task that requires such depth of knowledge, what, then, should we expect of the state of the art in ``AI?''  In this research, we document our experimental evaluation of the performance of OpenAI's \textsc{text-davinci-003} model, often-referred to as GPT-3.5, on the multistate multiple choice (MBE) section of the exam.  While we find no benefit in fine-tuning over GPT-3.5's zero-shot performance at the scale of our training data, we do find that hyperparameter optimization and prompt engineering positively impacted GPT-3.5's zero-shot performance.  For best prompt and parameters, GPT-3.5 achieves a headline correct rate of 50.3\% on a complete NCBE MBE practice exam, significantly in excess of the 25\% baseline guessing rate, and performs at a passing rate for both Evidence and Torts.  GPT-3.5's ranking of responses is also highly-correlated with correctness; its top two and top three choices are correct 71\% and 88\% of the time, respectively, indicating very strong non-entailment performance.  While our ability to interpret these results is limited by nascent scientific understanding of LLMs and the proprietary nature of GPT, we believe that these results strongly suggest that an LLM will pass the MBE component of the Bar Exam in the near future.



\section*{Introduction}
The legal system is becoming increasingly complex\cite{bib1}\cite{bib2}\cite{bib3}, leading to a need for technology to assist with the quantity, quality, and accessibility of legal services demanded by society. As in other domains, artificial intelligence and process engineering have promised help for decades to both non-professional and professional users of legal systems\cite{bib4}\cite{bib5}.  Significant research and development has gone into use cases like search and legal aid for laypeople, automated argumentation or brief construction, pre- and post-execution contract processes, due diligence and e-discovery, and judicial analysis\cite{bib6}\cite{bib7}\cite{bib8}.  However, the complexity of legal language and vastness of legal knowledge has made it historically difficult to develop systems that understand the nuances of legal tasks, and many systems have failed to deliver desired results or reach adoption.

As noted in \cite{bib9}, law is a field which is heavily reliant on the use of language, producing massive volumes of textual data \cite{bib10}. Documents such as briefs, memos, statutes, regulations, contracts, patents, and judicial decisions are continuously authored by lawyers, judges, and regulators \cite{bib2}.  To make matters even more difficult, legal language is notoriously complex; lawyers and other legal professionals undertake nearly a decade of education and professional training to understand and generate it.  

Why is this language so ``complex?''  Why do so many proficient users of natural languages struggle with contracts and laws, even in their native tongue, to the point that descriptors like ``legalese'' or ``lawyer speak'' have become common parlance?  The answer is likely two-fold.  First, for both technical and cultural reasons, the grammar of legal language is significantly different than the grammar of normal language, featuring both highly-stylized customs and pedantically-precise phrasing.  The resulting sentence structures are typically much larger and more complex than normal language, as the number of clauses and ``distance'' over which clauses are connected exceeds the working memory of both human and non-human readers.  Second, by the very nature of common law and precedent, legal language is full of semantic nuance and history.  Words like ``security'' that have common meaning in normal language often have different, context-specific meanings in legal language.  Many words that do not occur at all in normal language, like ``estoppel'' or ``indemnitor,'' occur regularly in legal corpora.  This semantic depth and breadth traditionally required systems that interact with legal text to embed a large amount of domain-specific knowledge.  Viewed from this perspective, legal education and training is required to teach humans to understand and produce this very particular type of language, and it is no surprise that traditional models in NLP struggled in general legal task assessments.

In recent years, however, developments in natural language processing and computing have led to significant advances in state of the art performance. Leveraging advances in neural network research\cite{bib11}\cite{bib12}, sophisticated efforts have been made to build quasi-semantic models. The age of neural NLP can be traced to \cite{bib13}, which has been followed by successive waves of embedding \cite{bib14} \cite{bib15} and transformer-based large language models (LLMs) \cite{bib16} \cite{bib17} \cite{bib18} \cite{bib19} \cite{bib20}. In particular, transformer architectures, first introduced in \cite{bib25}, have revolutionized machine learning research, and have been most successfully applied to text and image modalities. The most famous and accessible of these LLMs is OpenAI's family of Generative Pre-trained Transformer models, commonly referred to as GPT. 


 As a proprietary model in production for OpenAI's customers, there is no guarantee that previously-published academic literature is still accurate.  However, as of July 2020, OpenAI reported that GPT-3 was ``an autoregressive language model with 175 billion parameters'' featuring 96 layers trained with a batch size of 3.2M.  While these numbers may be difficult to contextualize, those who have trained their own models can easily appreciate the effort involved. Since then, OpenAI has also launched or published a number of derivative models, most notably InstructGPT-3 and Codex 12B.  Colloquially, these recent models are referred to by many, including OpenAI, as GPT-3.5.  More specifically, as described on OpenAI's website, ``GPT-3.5 series is a series of models that was trained on a blend of text and code from before Q4 2021.''  Our results in this publication are based on \textsc{text-davinci-003}, which is ``an improvement on \textsc{text-davinci-002}'', which is ``an InstructGPT model based on \textsc{code-davinci-002}'', which is ``a base model [...] for pure code-completion tasks.''
 
 GPT-3 and derivative models are not, however, directly available for use in frameworks like PyTorch or Tensorflow; for both commercial and ethical reasons, access to OpenAI models has historically only been available through OpenAI's API, which is designed both to accomplish specific customer tasks and to provide a layer of legal and ethical moderation.  As of this publication, OpenAI's APIs offers text completion, code completion, image generation, and embedding generation endpoints.  In recent weeks, OpenAI has also released a public-facing chatbot version of GPT-3.5 known as ChatGPT, which reportedly resulted in over 1M user signups within six days of release. 

While GPT-3.5 and ChatGPT have demonstrated previously-unseen performance on zero-shot or few-shot tasks, they are not domain-specific models.  As reported in \cite{bib17}, OpenAI's models are trained on a combination of curated CommonCrawl data and high-quality reference data that, if we consider The Pile V1 as reference, may have included some material from public legal sources.  However, given the complex nature of legal language and GPT-3.5's training on general task performance, it is an open question as to whether state-of-the-art LLMs like GPT-3.5 can succeed in legal task assessments, let alone zero- or few-shot tasks.  In order to evaluate this question, we decided to test GPT-3.5 on the multistate multiple choice section of the Bar Exam, known as the Multistate Bar Examination (MBE), using zero-shot prompts for the \textsc{text-davinci-003} text completion API. 

\section*{Data}
Professional licensure exams like ``the Bar Exam'' are common across professional fields, including not just law, but also medicine, dentistry, pharmacy, accounting, and engineering.  While each jurisdiction (e.g., state) in the United States is responsible for administering its own law licensure requirements, the National Conference of Bar Examiners (NCBE) is the organization responsible for designing most of the bar examination materials used across the United States. In this research, we follow colloquial convention and refer to the NCBE's standardized exam format as ``the Bar Exam'' or ``the Bar,'' while abstract exams that may vary across countries or states are referred to as bar exams in the indefinite.  

For the individual test-taker, such bar exams are the culmination of years of education as well as preparation specific to each exam component. Successful performance on these exams generally requires two things: (i) the acquisition of a large amount of accumulated theoretical knowledge (semantics) and (ii) the ability to understand and answer exam-specific questions that often feature unique syntax.  Prior attempts to develop systems to take bar exams around the world have yielded mixed results, with significant exam-specific training required to even achieve such performance \cite{bib21} \cite{bib22} \cite{bib23}.

For most test-takers, the Bar Exam represents the most significant single challenge of their academic careers.  In order to be eligible, the typical applicant is required to complete at least seven years of post-secondary education, including a four-year bachelors degree and successful completion of three years of study at an ABA-accredited law school. Following graduation from law school, most applicants also invest substantial amounts of time and money into post-graduation Bar preparation training\cite{bib24}. This additional preparation is intended to not only solidify one's legal knowledge, but also critically to teach the applicant how to understand and answer the exam's questions.  Despite the incredible effort of the average test-taker, approximately one out of every five still fails to pass the exam on their initial attempt.
\\
 As a historical matter, the specific components of the Bar exam once differed widely from state to state. Recently, however, most states have adopted the Uniform Bar Examination (UBE). The UBE features three components: (i) a multiple choice test, (ii) an essay test, and (iii) scenario-based performance test. The multiple choice component, referred to as the Multistate Bar Examination or MBE, is typically worth 50\% of an overall bar exam score. 

As the MBE is a single component of an exam, most jurisdictions do not require a minimum MBE score.  The MBE is also scaled by jurisdictions and the NCBE after each exam window; for example, a raw score of roughly $\sim60\%$ may yield an approximate scaled score of 133, which would be enough to pass in a significant number of jurisdictions, including New York, Illinois, 
 and the District of Columbia.

Questions on the MBE are designed to test both legal knowledge and reading comprehension skills, requiring above-average semantic and syntactic command of the English language.  Instead of posing direct legal questions as they might appear in a textbook or theory exam, most MBE questions present the test-taker with a fictional situation.  Descriptions of the facts are typically embellished with details; some of these details are critically important, while others are added only to distract or confuse the reader.  A public sample provided by the NCBE on their website is shown below:

\begin{verbnobox}[\small\slshape\hspace{.5in}]
Question: A man sued a railroad for personal injuries suffered when his
car was struck by a train at an unguarded crossing.  A major issue is
whether the train sounded its whistle before arriving at the crossing.
The railroad has offered the testimony of a resident who has lived near
the crossing for 15 years.  Although she was not present on the occasion
in question, she will testify that, whenever she is home, the train always 
sounds its whistle before arriving at the crossing.

Is the resident’s testimony admissible?

(A) No, due to the resident’s lack of personal knowledge regarding the
incident in question.
(B) No, because habit evidence is limited to the conduct of persons,
not businesses.
(C) Yes, as evidence of a routine practice.
(D) Yes, as a summary of her present sense impressions.
\end{verbnobox}

The MBE portion of the Bar consists of approximately 200 questions like the sample above.  As detailed in Table \ref{tab:student_performance}, real examinations present test-takers with 25 questions from eight categories, seven of which correspond to specific areas of law and one of which is used by the NCBE to experiment with test design.  In some instances, a subset of these these questions are removed from final scoring of an exam by state bars or the NCBE; both individual state bars and the NCBE assess the performance of test-takers within and across states, dropping some questions and scaling the raw scores to maintain consistency across jurisdictions.  As part of its role in exam design and preparation, the NCBE also maintains statistical information regarding exam performance. For comparison, we show their reported average accuracy of students by question category in Table \ref{tab:student_performance}.  In absolute terms, this table makes clear the difficulty of the exam, as the average student answers more than one in four questions incorrectly.

\begin{table}[tbp]
  \centering
  \small
  \begin{tabular}{rcc}
    \textbf{Question Category} & \textbf{Number of Questions} & \textbf{Correct Rate} \\
    \hline
    \\
    Torts & 25 & 71\%  \\
    Contracts & 25 & 70\% \\
    Evidence & 25 & 65\% \\
    Real Property & 25 & 65\% \\
    Civil Procedure & 25 & 59\%  \\
    Constitutional Law & 25 & 72\% \\
    Criminal Law and Procedure & 25 & 71\% \\
    Experimental Questions & 25 & N/A \\
    \\
    \thickhline
    \\
    & 200  & 68\%  \\
    & TOTAL  &  AVERAGE
    \\
    \\
  \end{tabular}
 \caption{\centering{NCBE-Reported Average Student Performance by Question Category}}
  \label{tab:student_performance}
\end{table}

For this research, we purchased the standard test preparation material offered by the NCBE, including practice questions and simulated exams for the MBE portion of the Bar Exam.  While we cannot redistribute these materials, researchers interested in replicating the results contained in this paper can purchase these data for approximately 300 USD directly from the NCBE's online store.  All reported task assessments are based on the practice exam and answer key available in the downloadable MBE Study Aid purchased in December 2022, dated in the document as of 2019.  The body of each question was automatically extracted with its four multiple choice options and stored separately from the answer key, which consisted solely of the correct letter answer for each question.  The answer key is found at the back of the document in a simple table, provided without explanations of correct and incorrect answers.  We specifically chose this exam PDF for task assessment instead of other simulated exams because the low probability that any prior model exposure to the document could have been learnable.

\section*{Methods}
As discussed above, our experimental evaluation of GPT-3.5 involved using zero-shot prompts for the \textsc{text-davinci-003} text completion API. In this section, we detail how we implemented this experiment, including the design and iteration of these prompts, related API hyperparameters, and an attempt at fine-tuning the mode.  While replication of this research requires access NCBE's material and an OpenAI account, we have done our best to provide researchers with as much detail as we have ourselves.

\subsubsection*{Prompt Engineering and Responses}
Our scientific understanding of large language models is nascent, and we often do not understand how or why they produce the outputs they do.  However, despite this scientific gap, we do know that LLMs are often highly sensitive to the prompts they are provided.  The ``art'' of crafting such prompts is typically referred to as ``prompt engineering,'' and details of prompt engineering are critical to replication of 
studies involving LLMs.  In this research, we experimented substantially with prompt engineering.  The following prompt types were tested:
\begin{enumerate}
 \item Single choice only: Ask the model for a single multiple choice answer only.
 \item Single choice and explanation: Ask the model for a single multiple choice answer with an explanation of its reasoning.
 \item Top two choices only: Ask the model for its best answer and a backup answer.
 \item Top two choices and explanation: Ask the model for its best and backup answer with an explanation of its reasoning.
 \item Top two choices and re-prompt: Ask the model for its best and backup answer, then re-prompt the model between these two choices, similar to the iterative ``strike-out'' heuristic that many human test-takers are taught.
 \item Rank order all choices: Ask the model to rank order all four multiple choice answers.
 \item Rank order top three choices: Ask the model to rank order its top three multiple choice answers.
\end{enumerate}

Results did not vary substantially between many of these prompts.  However, the last prompt strategy - rank-ordering of the top three choices - improved model correctness substantially.  Unfortunately, because we have no direct insight into the head layers of GPT-3.5, we have no ability to comment further on why this prompt variation impacted the model's behavior in ways that other prompts did not.  We speculate that this prompt best combined non-entailment performance, i.e., rejection of most incorrect answer, with probabilistic entailment and recall.  Below is an example of how this prompt manifested with a partially-redacted version of a real NCBE question:

\begin{verbnobox}[\small\slshape\hspace{.5in}]
Please answer the following Bar Exam question in the following rank
order format: 
First Choice: <LETTER>
Second Choice: <LETTER>
Third Choice: <LETTER>

Question: A plaintiff domiciled in State A has brought a federal 
diversity action in State A against a defendant domiciled in 
State B, [...]
(A) Move for discovery [...]
(B) Move for judgment on the pleadings, [...]
(C) Move for sanctions against the plaintiff [...]
(D) Move to dismiss the action for lack of personal jurisdiction[...]

Answer:
\end{verbnobox}

Upon querying the text completion API endpoint, we then received back responses like those below:
\begin{verbnobox}[\small\slshape\hspace{.5in}]
First Choice: D
Second Choice: B
Third Choice: A
\end{verbnobox}

The prompt and complete JSON response, including the OpenAI API request ID, were logged for all simulated exams.  Each line of the text completion response was parsed and stored for scoring or qualitative analysis.  In a small number of cases ($<1\%$), responses included natural language or format variations such as ``My first choice is (D)'' and these variations were handled through exception cases in our parser.  No responses were manually altered or evaluated by humans.

From a technical perspective, all of these prompts are related to traditional textual entailment tasks where a model must evaluate whether a statement is truthful or non-truthful.  In most extant research on the topic, this problem is formulated relative to another statement or body of knowledge, and tasks are assessed by independently evaluating single claims in a binary setting.  In our zero-shot exam simulation, unlike most extant research on entailment problems, we have little control over the framing of the hypothesis, claim, or body of knowledge.  We have no insight into any knowledge graphs or state models, explicit or implicit, that exist in GPT.  Furthermore, in some cases, multiple choices may be correct from an entailment perspective, and test-takers must rank order their choice based on knowledge of exam design.  As such, there are elements of this test that are more similar to search and relevancy scoring than simple binary entailment/non-entailment problems.

\subsection*{(Hyper)parameters for GPT-3}
The results of machine learning and computational research generally are often highly sensitive to model parameters or hyperparameters.  In this research, in addition to varying prompts as detailed above, we also evaluated how hyperparameters like model ``temperature'' impacted the performance of the model.  While our ability to interpret the nature or impact of these hyperparameters is limited by OpenAI documentation and API functionality, we evaluated the following parameters:
\begin{enumerate}
  \item \textsc{temperature}: Sampling temperature; 0.0 is deterministic, higher is more ``random.''  We tested values in $\{0.0, 0.25, 0.5, 0.75, 1.0\}$.
  \item \textsc{top\_p}: Nucleus sampling probability.  We tested values in $\{0.75, 1.0\}$.
  \item \textsc{best\_of}: ``Generates [N] completions server-side and returns the "best" (the one with the highest log probability per token).''  We tested values in $\{1, 2, 4\}$.
  \item \textsc{max\_tokens}: Maximum number of tokens to generate.  For prompts without an explanation, we tested values in $\{16, 32\}$.  For prompts with an explanation, we tested values in $\{128, 256, 1024\}$.
\end{enumerate}

\subsection*{Fine-tuning}
LLMs like GPT-3.5 have received so much interest in part because their zero-shot or few-shot performance is so good.  Despite this, in some circumstances, subsequent supervised or unsupervised re-training of some or all layers of an LLM may improve performance \cite{bib26}\cite{bib27}.  OpenAI does make some retraining or ``fine-tuning'' capabilities available through its API, and these API endpoints do allow for some control of the training process like learning rates or batch sizes.  We did attempt to fine tune \textsc{text-davinci-003} by providing it with 200 unseen, simulated MBE bar exam questions with correct and incorrect explanations.  We provided the training samples both with and without explanatory text from the answer guide.  In total, we trained six fine-tuned models, altering training prompts, training responses, batch size, learning rate, and prompt weighting.  However, in all cases, the fine-tuned model significantly underperformed \textsc{text-davinci-003} itself.  Due to the scarcity of high-quality data for training and assessment, we did not pursue fine-tuning of GPT models further, and these results possibly confirm LLM fine-tuning risks observed by others \cite{bib28}.

\section*{Results}
\begin{figure}[tbp]
    \centering
    \includegraphics[width=5in]{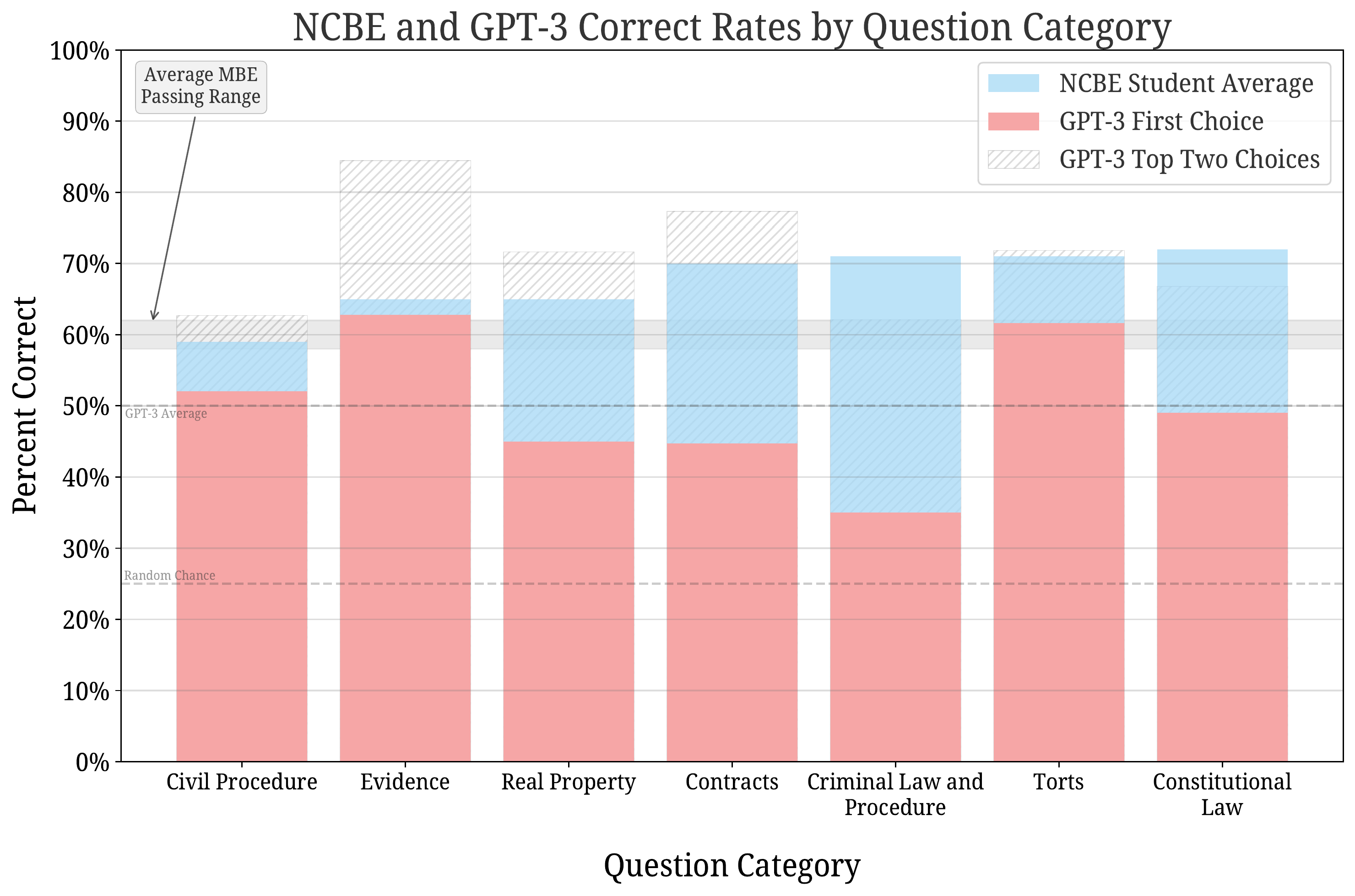}
    \caption{Summary of performance by question category for GPT-3.5 and NCBE-Reported Students}
    \label{fig:accuracy_bar_chart}
\end{figure}

In total, we executed 107 sample exams across the prompts and parameter values described above.   Out of these prompts, prompt style \#7 - rank-ordering of the top three choices - performed best, and we collected 41 sample runs across parameter combinations for this prompt.  The performance of these runs is summarized in Figure \ref{fig:accuracy_bar_chart} and Table \ref{tab:gpt_ncbe_performance}, including a comparison with baseline student and passing rates.\footnote{As the MBE is just one component of the overall bar exam, students cannot ``pass'' the exam solely by achieving 58-62\% on the multiple choice; however, across a plurality of states, this score, in combination with adequate performance on other components, produces a satisfactory result.\\\\Additional tables summarizing variation within and across hyperparameters is provided in the Supplementary Information section.}

\begin{table}[tbp]
  \centering
  \small
  \begin{tabular}{rcccc}
        & \textbf{GPT} & \textbf{GPT Top 2} & \textbf{GPT Top 3} & \textbf{NCBE} \\
        \hline
        \\
        Evidence & 63\% & 84\% & 98\% & 65\% \\
        Torts & 62\% & 72\% & 93\% & 71\% \\
        Civil Procedure & 52\% & 63\% & 79\% & 59\% \\
        Constitutional Law & 49\% & 67\% & 87\% & 72\% \\
        Real Property & 45\% & 72\% & 85\% & 65\% \\
        Contracts & 45\% & 77\% & 86\% & 70\% \\
        Criminal Law \& Procedure & 35\% & 62\% & 86\% & 71\% \\
        \\
        \hline
        \\
        AVERAGE & 50\% & 71\% & 88\% & 68\%
        \\
    \end{tabular}
    \caption{Summary of performance by question category for GPT-3.5 and NCBE-Reported Students}
    \label{tab:gpt_ncbe_performance}
\end{table}

The table and figure clearly show that GPT is not yet passing the overall multiple choice exam.  However, GPT is significantly exceeding the baseline random chance rate of 25\%.  Furthermore, GPT has reached the average passing rate for at least two categories, Evidence and Torts.

On average across all categories, GPT is trailing human test-takers by approximately 17\%.  In the case of Evidence, Torts, and Civil Procedure, this gap is negligible or in the single digits; at 1.5 times the standard error of the mean across our test runs, GPT is already at parity with humans for Evidence questions.  However, for the remaining categories of Constitutional Law, Real Property, Contracts, and Criminal Law, the gap is much more material, rising as high as 36\% in the case of Criminal Law.  

This performance gap may be attributable to at least two issues.  First, it is possible that GPT's poor performance corresponds to bodies of knowledge that were absent from its training data or removed during subsequent model compression or fine-tuning.  Our ability to speculate further is limited by lack of information about GPT-3.5's original provenance or subsequent architecture or re-training changes.  Second, it is possible that GPT's poor performance on these categories is a result of the complex or purposefully-confusing language used by the exam's designers.  

In order to explore these two possibilities, we next examine how ``close'' GPT is to correct.  If GPT truly lacks knowledge about an area of law, then we should expect it to have low correlation between the rank of its answers and correctness.  If, on the other hand, its second or third best choices are very often correct, then we can infer that the design of the questions may be responsible for poor performance.  As shown by Table \ref{tab:student_performance}, some sections of the Bar are ``trickier'' than others, and so this finding may itself confirm what is held to be common knowledge by many human test-takers.

To understand this rank order performance, Figure \ref{fig:accuracy_bar_chart}  and Table \ref{tab:gpt_ncbe_performance} also include information about the performance of the model including its second-best and third-best answers.  As shown by the gray dashed region in the figure and the ``GPT Top 2'' column in the table, GPT's second best answer is highly correlated with correctness.  In all categories, the top two answers exceed the baseline random chance rate of 50\%, and in five out of seven categories, exceed the NCBE-reported averages.  The table also includes a summary of the top three GPT answer performance in the ``GPT Top 3'' column, which similarly shows strong overall correlation.  Except for Civil Procedure, which is notably also the worst category for human test-takers, GPT's answers significantly exceed the baseline random chance rate.

\section*{Conclusion and Future Work}
In this research, we document our experimental evaluation of GPT-3.5 on the MBE portion of NCBE's model Bar Exam.  Across all prompts and hyperparameter values, GPT-3.5 significantly outperformed the baseline rate of random guessing.  Without any fine-tuning, it currently achieves a passing rate on two categories of the Bar and achieves parity with human test-takers on one.  Its rank-ordering of possible choices is strongly correlated with correctness in excess of random chance, confirming its general understanding of the legal domain.

Overall, we find that GPT-3.5 significantly exceeds our expectations for performance on this task.  Despite thousands of hours on related tasks over the last two decades between the authors, we did not expect GPT-3.5 to demonstrate such proficiency in a zero-shot settings with minimal modeling and optimization effort.  While our ability to interpret how or why GPT-3.5 chooses between candidate answers is limited by understanding of LLMs and the proprietary nature of GPT, the history of similar problems strongly suggests that an LLM may soon pass the Bar.  Based on anecdotal evidence related to GPT-4 and LAION's Bloom family of models, it is quite possible that this will occur within the next 0-18 months.  

Many of the outstanding questions or improvements on this problem require either collaboration with OpenAI or the use of an alternative model that can be directly inspected, such as those maintained by EleutherAI, BigScience, or LAION.  We intend to replicate our experimental design and continue fine-tuning with models from the GPT-J, GPT-Neo, and Bloom families.  Separately, as noted above, the MBE is just one component of the overall Bar exam; we intend to assess both GPT-3.5 and other models mentioned above on both the essay (MEE) and situational performance (MPT) sections of the Exam in future work.

\section*{Acknowledgments}
Although the original draft of this paper was written by the authors, portions of this paper were fine-tuned by \textsc{text-davinci-003} for clarity, which only struggled lightly with the bibtex citations and formatting.

\nolinenumbers

%
%
%

\pagebreak

\section*{Supplementary Information}
As of this preprint, the Supplementary Information is available on GitHub at the following URL:
\\
\url{https://github.com/mjbommar/gpt-takes-the-bar-exam}.

\begin{table}[h]
    \centering
    \small
    \begin{tabular}{|r|c|c|c|c|}
        \textbf{Temperature} & \textbf{GPT} & \textbf{GPT Top 2} & \textbf{GPT Top 3} & \textbf{Samples} \\
        0.0 & 49.86\% & 71.77\% & 89.00\% & 5 \\
        0.5 & 50.19\% & 71.05\% & 88.20\% & 18 \\
        1.0 & 49.79\% & 70.65\% & 86.95\% & 18 \\
    \end{tabular}
    \caption{Summary of model performance by GPT \textsc{temperature} parameter}
    \label{tab:correct_by_temperature}
\end{table}
    
\begin{table}[h]
    \centering
    \small
    \begin{tabular}{|r|c|c|c|c|}
        \textbf{Best Of} & \textbf{GPT} & \textbf{GPT Top 2} & \textbf{GPT Top 3} & \textbf{Samples} \\
        1 & 49.51\% & 70.59\% & 87.27\% & 15 \\
        2 & 50.27\% & 71.22\% & 88.17\% & 14 \\
        4 & 50.20\% & 71.13\% & 87.84\% & 12 \\
    \end{tabular}
    \caption{Summary of model performance by GPT \textsc{best\_of} parameter}
    \label{tab:correct_by_bestof}
\end{table}

\begin{figure}[h!]
    \includegraphics[width=5.5in]{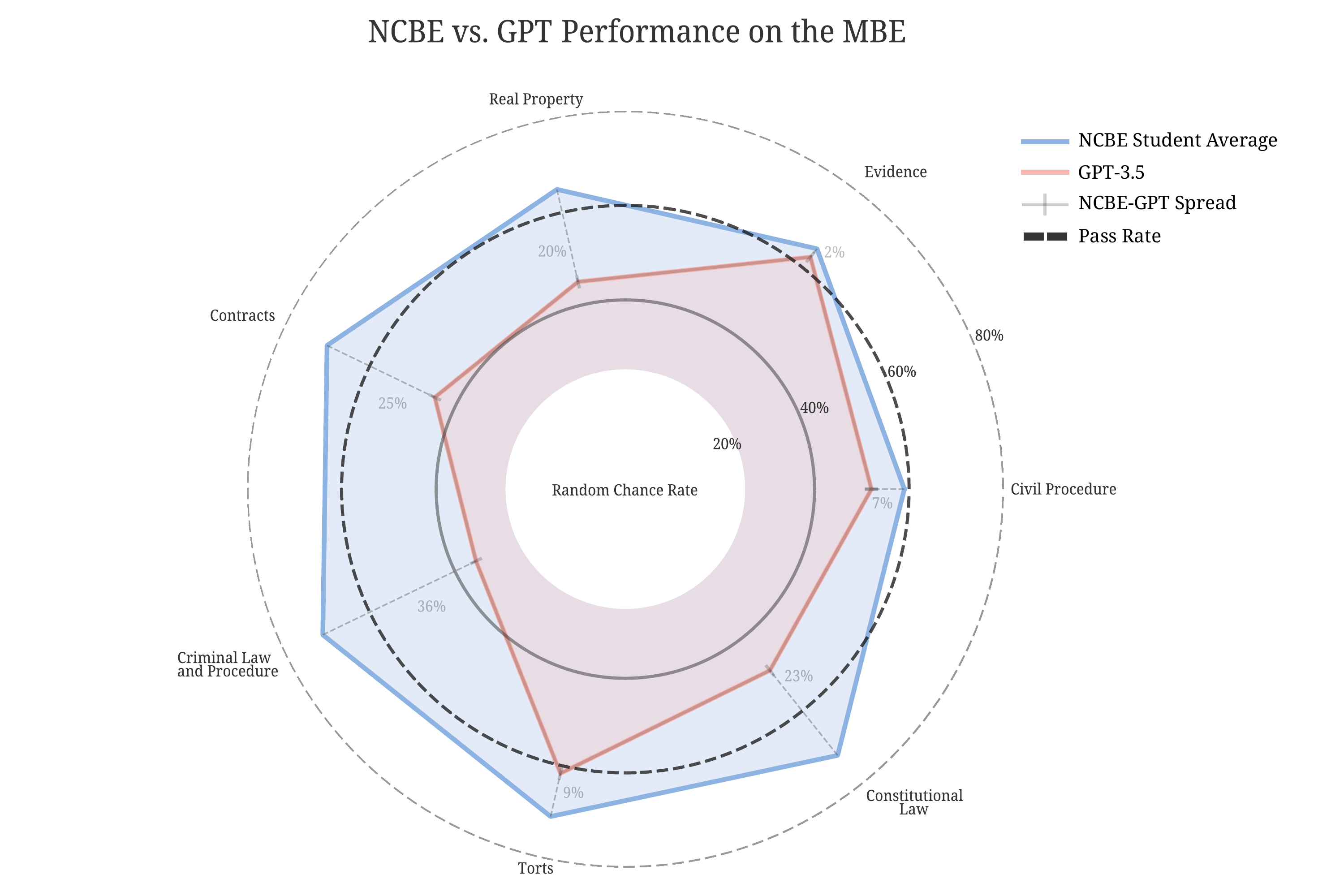}
    \caption{Accuracy by Question Category for GPT and Average Test-Takers}
    \label{fig:accuracy_radar_chart}
\end{figure}

\end{document}